\documentclass[10pt,twocolumn,letterpaper]{article}

\usepackage{cvpr}              
\usepackage{xcolor}
\usepackage{colortbl}
\usepackage{multirow}
\usepackage[accsupp]{axessibility} 

\definecolor{cvprblue}{rgb}{0.21,0.49,0.74}
\usepackage[pagebackref,breaklinks,colorlinks,citecolor=cvprblue]{hyperref}

\def\paperID{5091} 
\def\confName{CVPR}
\def\confYear{2024}

\title{Siamese Learning with Joint Alignment and Regression for \\ Weakly-Supervised Video Paragraph Grounding \vspace{-0mm}}
\author{
Chaolei Tan$^1$
\qquad Jianhuang Lai$^{1,2,3}$
\qquad Wei-Shi Zheng$^{1,2,3}$
\qquad Jian-Fang Hu$^{1,2,3}$\thanks{Corresponding author} \\
$^1${School of Computer Science and Engineering, Sun Yat-sen University, China}\\
$^2${Guangdong Province Key Laboratory of Information Security Technology, China}\\
$^3${Key Laboratory of Machine Intelligence and Advanced Computing, Ministry of Education, China}\\
\tt \small tanchlei@mail2.sysu.edu.cn, stsljh@mail.sysu.edu.cn, wszheng@ieee.org, hujf5@mail.sysu.edu.cn \vspace{-7.5mm}}

\begin{document}
\maketitle
\begin{abstract}
Video Paragraph Grounding (VPG) is an emerging task in video-language understanding, which aims at localizing multiple sentences with semantic relations and temporal order from an untrimmed video. However, existing VPG approaches are heavily reliant on a considerable number of temporal labels that are laborious and  time-consuming to acquire. In this work, we introduce and explore Weakly-Supervised Video Paragraph Grounding (WSVPG) to eliminate the need of temporal annotations. Different from previous weakly-supervised grounding frameworks based on multiple instance learning or reconstruction learning for two-stage candidate ranking, we propose a novel siamese learning framework that jointly learns the cross-modal feature alignment and temporal coordinate regression without timestamp labels to achieve concise one-stage localization for WSVPG. Specifically, we devise a Siamese Grounding TRansformer (SiamGTR) consisting of two weight-sharing branches for learning complementary supervision. An Augmentation Branch is utilized for directly regressing the temporal boundaries of a complete paragraph within a pseudo video, and an Inference Branch is designed to capture the order-guided feature correspondence for localizing multiple sentences in a normal video. We demonstrate by extensive experiments that our paradigm has superior practicability and flexibility to achieve efficient weakly-supervised or semi-supervised learning, outperforming state-of-the-art methods trained with the same or stronger supervision.
\end{abstract}
\vspace{-5mm}
\section{Introduction}
\label{sec:intro}
\begin{figure}[tp]
    \centering
    \includegraphics[width=1\linewidth]{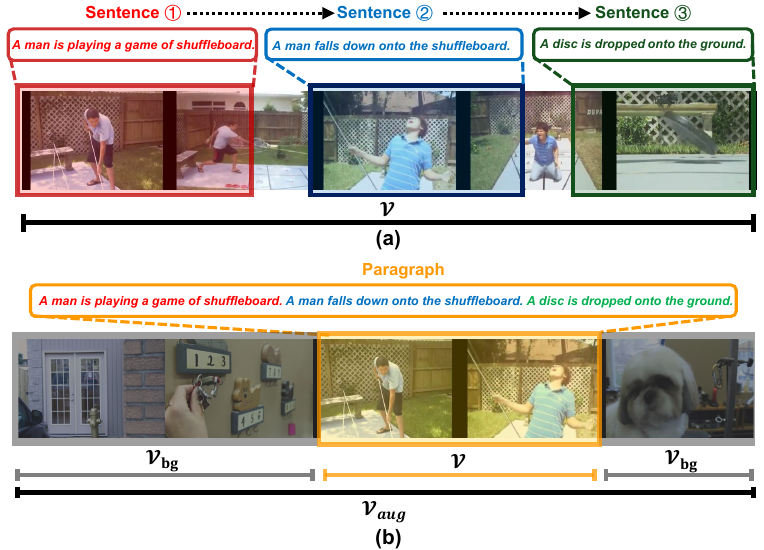}
    \vspace{-7.5mm}
    \caption{(a) Chronological cross-modal alignment in a video and its paired sentences. (b) Pseudo boundary supervision for regressing paragraph timestamps in a composed pseudo video.}
    \vspace{-6mm}
    \label{fig: intro}
\end{figure}
Natural Language Video Grounding (NLVG) is an essential area in vision-language understanding, which has received increasing attention due to its wide range of real-world applications such as video retrieval~\cite{vr_1, vr_2, vr_3, vr_4, vr_5, vr_6}, video summarization~\cite{vidsum_1, vidsum_2, vidsum_3, vidsum_4, vidsum_5, vidsum_6}, action segmentation~\cite{action_seg_1, action_seg_2, action_seg_3, action_seg_4, action_seg_5, action_seg_6}, video question answering~\cite{vqa_1, vqa_2, vqa_3, vqa_4, vqa_5, vqa_6}, etc. Most previous works focus on tackling Video Sentence Grounding (VSG) proposed in ~\cite{tall, didemo}, which targets at localizing the temporal boundaries of an individual sentence from an untrimmed video. However, localizing single sentences can be ambiguous because the contextual information conveyed by multiple sentences is necessary to uniquely determine the temporal locations of input queries. To alleviate such issue, Bao \etal~\cite{dense_grounding} proposed to contextualize video grounding into localizing multiple events indicated by sentences of a paragraph from the video, which is called Video Paragraph Grounding (VPG).

Although remarkable progress has been attained in tackling VSG and VPG problems under a fully-supervised setting, the extremely prohibitive overheads of manually annotating temporal boundaries for language queries limit these methods to utilize large-scale video-text data. In addition, the subjectivity of annotators inevitably brings label noises that may adversely affect model training. Recently, weakly-supervised methods free of temporal annotations have become increasingly popular in the area of video grounding, aiming to address the above limitations. These approaches can be mainly categorized into multiple instance learning methods and reconstruction learning methods and are typically based on a propose-and-rank pipeline. Nevertheless, both of these paradigms assume the contributions of proposals to the contrastive or reconstruction loss accurately represent the proposal quality, which is not necessarily the case during model learning. Moreover, the quadratic complexity of proposal schemes prevents them from scaling up in parameters or training data, and the adopted supervision of video-level contrast or query-based reconstruction in prior works is not temporal-sensitive thus suffering from a huge gap with fully-supervised temporal guidance. Besides, all of the weakly-supervised methods in video grounding are tailored for tackling VSG while the weakly-supervised setting of VPG (i.e., WSVPG) has been understudied so far.

To circumvent the aforementioned drawbacks and explore an efficient weakly-supervised framework for VPG, we seek to mine the unique characteristics and underlying supervision from the intrinsic structure of video-paragraph pairs for model training. On the one hand, as observed in Figure~\ref{fig: intro} (a), the temporal location of an event is highly correlated with the position of the sentence describing that event in the paragraph. For example, the sentence appearing in the middle of the paragraph tends to have stronger relevance with visual content located around the temporal midpoint of the video. On the other hand, dense visual events mentioned in the paragraph approximately represent the global video content that unambiguously distinguishes itself from another video, which can be observed in Figure~\ref{fig: intro} (b). Therefore, inserting the query-related video into another irrelevant video automatically generates pseudo boundary labels close to the ground-truth when regarding the complete paragraph as language query for video grounding.

Motivated by the above observation, we propose a novel Siamese Grounding TRansformer (SiamGTR) for WSVPG. It jointly learns the cross-modal alignment and boundary regression via two siamese branches without generating proposals. Specifically, we propose to construct an Augmentation Branch (AB) which takes as input a pseudo video and adopts a complete paragraph as the language query to learn high-quality boundary supervision for localization. Also, an Inference Branch (IB) is designed to receive a normal video as input and is enforced to capture the order-guided cross-modal correspondence for attending over the specific video content relevant to each sentence. The two weight-sharing branches are effective to transfer complementary supervision for joint boundary prediction and feature association, which yields a weakly-supervised model with superior generalization through a concise one-stage pipeline. Extensive experiments verify the effectiveness of our model and show our method with the same or weaker supervision surpasses prior state-of-the-arts. In summary, our contributions are:
\begin{itemize}
    \item We introduce the task of Weakly-Supervised Video Paragraph Grounding (WSVPG), which aims to train a model for localizing multiple events indicated by queries without the supervision of timestamp labels.
    \item We propose a novel Siamese Grounding TRansformer (SiamGTR) for concise and efficient one-stage weakly-supervised learning of video paragraph grounding. It is composed of two weight-sharing branches including an Augmentation Branch (AB) for learning boundary regression of pseudo boundaries and an Inference Branch (IB) for learning order-guided cross-modal feature alignment.
    \item Extensive experiments verify the efficacy of our method, and demonstrate that our framework under the same or weaker supervision outperforms state-of-the-arts.
\end{itemize}

\section{Related Work}
\label{sec:related}
\vspace{-1mm}
\subsection{Video Sentence Grounding}
\vspace{-1mm}
\noindent\textbf{Fully-Supervised Video Sentence Grounding.} Plenty of approaches~\cite{tall, didemo, man, ablr, debug, scdm, gdp, tsp-prl, 2dtan, lgi, drn, vslnet, 2dtan_pami, vslnet_pami, cbp, qspn, vlgnet, bpnet, gtr, smin, matn, mmn, lpnet, slp, mgsl, g2l, 2dtvg, where2focus} have been proposed to address Fully-Supervised Video Sentence Grounding (FSVSG). In general, these works can be roughly categorized into proposal-based and proposal-free methods. Specifically, proposal-based methods~\cite{tall, didemo, man, scdm, 2dtan, drn, 2dtan_pami, cbp, qspn, bpnet, smin, mmn, slp, mgsl, g2l} involve a proposal generation stage using sliding windows~\cite{tall}, anchor proposals~\cite{man, gdp, scdm, debug, drn, cbp, slp, mgsl} or 2D temporal maps~\cite{2dtan, bpnet, 2dtan_pami, vlgnet, smin, mmn}, after which the generated proposals are ranked according to the query matching scores with potential post-processing like Non-Maximum Suppression (NMS). In contrast, proposal-free methods~\cite{ablr, tsp-prl, lgi, vslnet, vslnet_pami, gtr, matn, 2dtvg} remove the dense proposal generation and score ranking process by directly regressing timestamps~\cite{ablr, lgi, lpnet, gtr, 2dtvg}, predicting boundary distributions~\cite{vslnet, vslnet_pami, matn} or using reinforcement learning~\cite{tsp-prl}, which improves the computation efficiency and scenario adaptability. Following the line of proposal-free works, we propose a novel weakly-supervised regression-based framework that shows superior performance and practicability.

\noindent\textbf{Weakly-Supervised Video Sentence Grounding.} Weakly-Supervised Video Sentence Grounding (WSVSG)~\cite{tga, scn, vlanet, wstan, san, logan, marn, crm, vca, lcnet, rtptn, ccl, cnm, compose_wsvsg, cpl, iron, wstsg_uncertainty} has become a popular research area because of the severe dependence of FSVSG approaches on laborious and expensive manual temporal annotations. Most of existing WSVSG methods are based on a two-stage pipeline using multiple instance learning~\cite{vlanet, wstan, lcnet, crm}, reconstruction learning~\cite{scn}, or the combination of both~\cite{cnm, cpl}. In particular, Chen \etal~\cite{compose_wsvsg} have proposed a video composition strategy to generate pseudo temporal labels for WSVSG, which is the most related work to ours. However, several inherent drawbacks are involved in this approach. Firstly, individual sentences only describe local video content, thus viewing the starting/ending locations of foreground video as temporal boundaries of an individual sentence produces a weak and noisy temporal alignment, which is unsuitable for accurate boundary supervision. Moreover, it simply adapts an existing fully-supervised proposal-based framework for weakly-supervised training, which leads to inferior generalization caused by the large train-test discrepancy. Distinct from all of the above works, we design a novel siamese framework to capture the essential characteristics of video paragraph grounding for weakly-supervised learning.
\vspace{-1mm}
\subsection{Video Paragraph Grounding}
\vspace{-1mm}
Video Paragraph Grounding (VPG) is introduced by Bao \etal ~\cite{dense_grounding}, which aims to jointly localize multiple sentences of a paragraph from an untrimmed video. Shi \etal~\cite{prvg} presented an end-to-end network by re-purposing transformers into language-conditioned regressors. Jiang \etal~\cite{svptr} proposed to employ contrastive encoders for contrastive learning between video-paragraph pairs. Tan \etal~\cite{hscnet} proposed a hierarchical semantic correspondence network for modeling hierarchical video-language alignment and grounding multiple levels of language queries in the video. Particularly, Jiang \etal~\cite{svptr} first explored Semi-Supervised Video Paragraph Grounding (SSVPG) to relieve the annotation burden of temporal labels. However, SSVPG is still not quite practical considering the expensive cost of temporal annotations in untrimmed videos. Besides, these semi-supervised methods still require a considerable proportion of temporal labels up to at least $10\%$~\cite{svptr} for training. To thoroughly get rid of the temporal annotations, we pioneer to explore the weakly-supervised setting in VPG.
\vspace{-1mm}
\subsection{Siamese Networks}
\vspace{-1mm}
Siamese networks~\cite{siamese} are weight-sharing neural networks. There have been a wide range of scenarios where siamese networks are applied for achieving different purposes, such as face verification~\cite{deepface}, image recognition~\cite{siam_recognition}, object tracking~\cite{siam_tracking}, etc. In particular, siamese networks are commonly used in contrastive self-supervised learning methods~\cite{siam_unsup, simclr, simclrv2, dino, moco, simsiam, ibot, masked_siam, siameseIM, siammae}, in which augmented views of the same or different instance are forwarded into multiple weight-sharing network copies for learning a generalizable visual representation via instance-level discrimination. In this work, we explore a new way to combine the transferability of siamese architectures and the flexibility of transformer architectures for concise and efficient weakly-supervised learning of video paragraph grounding.

\section{Methodology}
\label{sec:method}
\subsection{Overview}
\begin{figure}[t]
    \centering
    \includegraphics[width=0.75\linewidth]{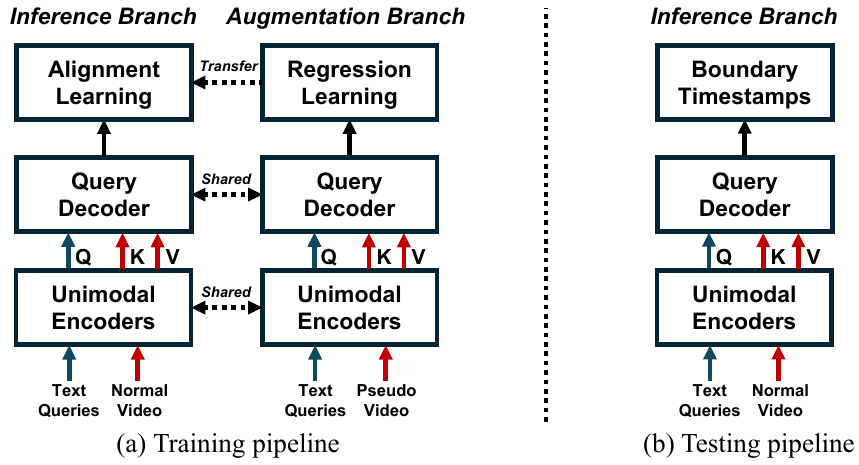}
    \vspace{-3mm}
    \caption{\footnotesize{Our siamese framework for joint alignment and regression.}}
    \label{fig: overview}
    \vspace{-6mm}
\end{figure}
\begin{figure*}[htbp]
    \centering
    \includegraphics[width=1\linewidth]{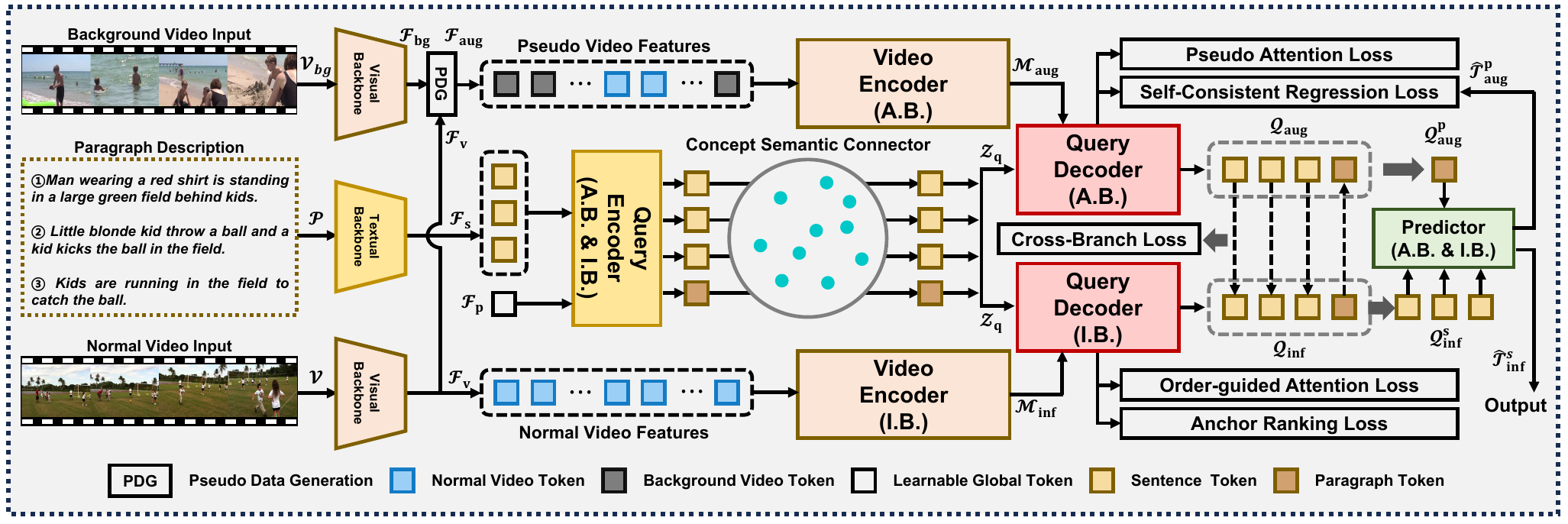}
    \vspace{-7.5mm}
    \caption{Illustration of the proposed Siamese Grounding TRansformer (SiamGTR) architecture. The augmentation branch (abbreviated as A.B.) takes the pseudo video features derived from randomly inserting the query-related video features into irrelevant video features. It learns to temporally regress the interval of interest from the pseudo video with the paragraph as query. The Inference Branch (abbreviated as I.B.) receives normal video features for learning the cross-modal feature alignment among multiple sentences in the video.}
    \label{fig: arch}
    \vspace{-5mm}
\end{figure*}
\noindent\textbf{Task Formulation.} Given an untrimmed video $\mathcal{V}$ and a paragraph query $\mathcal{P}$ consisting of $N$ temporally ordered sentences $\{\mathcal{S}_i\}_{i=1}^{N}$, the goal of Video Paragraph Grounding (VPG) is to simultaneously localize the temporal intervals $\mathcal{T}=\{(\tau_i^{\text{st}}, \tau_i^{\text{ed}})\}_{i=1}^{N}$ of all the events described by sentences in the paragraph, where $\tau_i^{\text{st}}$ and $\tau_i^{\text{ed}}$ are the starting and ending timestamps for the $i$-th sentence $\mathcal{S}_i$, respectively.

\noindent\textbf{Siamese Learning.} An overview of our siamese framework is shown in Figure~\ref{fig: overview}. Overall, we jointly train an augmentation branch and an inference branch with shared parameters and structures for learning the complementary abilities of cross-modal feature alignment and temporal boundary regression. These two branches follow the same workflow to first encode the text queries and input video into unimodal features, after which the query features are iteratively used in the transformer decoder to extract relevant information from the video features for timestamp decoding. For testing, we only keep the pipeline of inference branch for boundary prediction. More architectural details are illustrated in Figure~\ref{fig: arch} and are elaborated in the following sections.

\subsection{Feature Extraction}
\noindent\textbf{Video Feature Extraction.} For each input video, we divide it into consecutive clips consisting of a fixed number of frames for feature extraction. Specifically, a frozen pre-trained 3D Convolutional Neural Network (3D-CNN)~\cite{c3d} and a linear projection layer are successively employed to obtain a 1D feature vector for each short clip, resulting in a video feature sequence $\mathcal{F}_\text{v} \in \mathbb{R}^{L\times D}$, where $L$ and $D$ are the sequence length and hidden dimension, respectively.

\noindent\textbf{Text Feature Extraction.} For each input paragraph consisting of $N$ sentences, we first utilize a frozen pre-trained word embedding model to tokenize and embed the text into a sequence of word vectors. Then, a bidirectional Gated Recurrent Unit (GRU)~\cite{gru} is employed on each sentence, and the last hidden states in both directions are concatenated and then projected by a linear layer to construct the sentence features $\mathcal{F}_\text{s}\in \mathbb{R}^{N\times D}$, where $D$ is the hidden dimension.

\subsection{Augmentation Branch}
The augmentation branch aims to learn accurate boundary regression from pseudo videos with paragraph queries, which naturally transfers to the inference branch via the shared feature space established by the siamese structure.

\noindent\textbf{Pseudo Data Generation.} To drive end-to-end weakly-supervised regression learning, the input stream of the augmentation branch should provide reliable and direct boundary supervision. Drawing inspiration from the boundary-sensitive video pretext tasks~\cite{bsp, bassl, compose_wsvsg}, we propose to utilize the synthesized videos paired with paragraph queries to serve as a well-suited source of surrogate boundary supervision. Specifically, for each input video $\mathcal{V}$ and its paragraph description $\mathcal{P}$, we randomly sample a background video $\mathcal{V}_{\text{bg}}$ to obtain irrelevant video content to $\mathcal{P}$. Because $\mathcal{P}$ could be treated as a unique referring expression specific to $\mathcal{V}$, the search space of background videos can be the entire training set for increasing the data diversity. Denote video features of $\mathcal{V}$ and $\mathcal{V}_{\text{bg}}$ as $\mathcal{F}_\text{v}$ and $\mathcal{F}_\text{bg}$ respectively, we construct the pseudo video features $\mathcal{F}_\text{aug}$ as:
\vspace{-1mm}
\begin{equation}
    \mathcal{F}_\text{aug} = \text{NRS}\left(\text{Concat}\left(\mathcal{F}_\text{bg}^{1:I}, \text{RRS}(\mathcal{F}_{\text{v}}), \mathcal{F}_\text{bg}^{I+1:}\right)\right)
    \vspace{-1mm}
\end{equation}
where $I$ is a randomly generated inserting index within the feature sequence $\mathcal{F}_{\text{bg}}$ and $\text{Concat}\left(\cdot\right)$ denotes the temporal concatenation. $\text{RRS}\left(\cdot\right)$ and $\text{NRS}\left(\cdot\right)$ respectively represent a Random Re-Sampling operation to stochastically re-scale the length of $\mathcal{F}_\text{v}$ and a Normalized Re-Sampling operation that converts the length of $\mathcal{F}_\text{aug}$ into a fixed number of $T$. Thereafter, we compute the synthesized temporal boundaries $(\tau_{\text{aug}}^{\text{st}}, \tau_{\text{aug}}^{\text{ed}})$ in the pseudo video $\mathcal{V}_\text{aug}$ as follows:
\vspace{-1mm}
\begin{equation}
    \tau_{\text{aug}}^\text{st} = \frac{I + \Delta I^\text{st}}{rL + L_{\text{bg}}}, \space\space\space\space \tau_{\text{aug}}^\text{ed} = \frac{I + rL - \Delta I^\text{ed}}{rL + L_{\text{bg}}}
    \vspace{-1mm}
\end{equation}
where $L$ and $L_{\text{bg}}$ respectively indicate the length of $\mathcal{F}_\text{v}$ and $\mathcal{F}_{\text{bg}}$ with $r$ being the random re-scaling factor of $\text{RRS}\left(\cdot\right)$. To alleviate potential synthesis artifacts and boundary uncertainty, we further introduce a Random Boundary Shifting (\text{RBS}) strategy that incorporates small random offsets $\Delta I^\text{st}$ and $\Delta I^\text{ed}$ into computing the pseudo labels, which is simple yet effective to boost the quality of boundary supervision.

\noindent\textbf{Video Encoder.} Based on the obtained pseudo video features $\mathcal{F}_\text{aug}$, we then encode the temporal contextual information across multiple clips by feeding $\mathcal{F}_\text{aug}$ to a video encoder, which follows a similar architecture of DETR encoders~\cite{attention, detr, dab-detr}, i.e., each encoder layer consists of a multi-head self-attention module equipped with additive sinusoidal positional encodings and a feed-forward network. For simplicity, we omit the layer index and denote the input of each video encoder layer as $\mathcal{X}_\text{aug}$, then a set of Modulated Positional Encodings (MPE) are given as:
\vspace{-1mm}
\begin{equation}
    \mathcal{X}_{\text{pe}} = \text{MLP}\left(\mathcal{X}_\text{aug}\right) \odot \text{PE}\left(\mathcal{X}_\text{aug}\right)
    \vspace{-1mm}
\end{equation}
where $\mathcal{X}_{\text{pe}}\in \mathbb{R}^{T\times D}$ and $\odot$ denotes element-wise multiplication. $\text{PE}\left(\cdot\right)$ is the sinusoidal function~\cite{attention} with respect to temporal locations of $\mathcal{X}_\text{aug}$, $\text{MLP}\left(\cdot\right)$ denotes a two-layer feed-forward network computed on $\mathcal{X}_\text{aug}$. In each encoder layer, $\mathcal{X}_{\text{pe}}$ is only added to the input of projection layers of queries and keys in the self-attention mechanism. The encoded video features of the last video encoder layer, also called the encoder memory features $\mathcal{M}_\text{aug}$, are further fed to the query-guided decoder for localization prediction.

\noindent\textbf{Query Encoder.} Since there are strong contextual correlations among multiple sentences in the paragraph query, here we utilize a vanilla transformer encoder~\cite{attention} to help reason the semantic and chronological relationships of events. To extract a global representation of the complete paragraph, we initialize an extra learnable query token $\mathcal{F}_{\text{p}}$ and integrate it with the sentence features $\mathcal{F}_\text{s}$ by concatenation. Thus, the input of the query encoder is constructed as follows:
\vspace{-1mm}
\begin{equation}
    \mathcal{X}_{\text{q}} = \left[\mathcal{F}_{\text{p}}, \mathcal{F}_\text{s}^{1}, \mathcal{F}_\text{s}^{2}, \ldots, \mathcal{F}_\text{s}^{i}, \ldots \mathcal{F}_\text{s}^{N}\right]
    \vspace{-1mm}
\end{equation}
where $\mathcal{X}_{\text{q}}\in \mathbb{R}^{(N+1)\times D}$ and $\mathcal{F}_\text{s}^{i}$ is the $i$-th sentence feature in $\mathcal{F}_\text{s}$. We first normalize $\mathcal{X}_{\text{q}}$, add fixed sinusoidal positional encodings to it, and then iteratively employ self-attention modules and feed-forward networks to contextualize the global and local tokens that respectively represent the query semantics of the entire paragraph and the sentences. The output features of the last query encoder layer are denoted as $\mathcal{Z}_\text{q}$ and are further forwarded to the query decoder.

\noindent\textbf{Conceptual Semantic Connector.} Although the siamese network structure can implicitly transfer boundary knowledge from the augmentation branch to the inference branch, there is still a certain semantic gap between the query representations of short sentences and long paragraphs. To narrow this gap, we develop a Conceptual Semantic Connector (CSC) module for explicit semantic guidance. Specifically, we first collect high-frequency linguistic concepts, including verbs and nouns from the training corpus, and then construct a set of dictionary features by selecting and projecting the Glove vectors~\cite{glove}. The loss $\mathcal{L}_\text{csc}$ is computed as:
\vspace{-1.5mm}
\begin{equation}
    \mathcal{L}_\text{csc} = \text{BCE}\left(y_\text{cept}^\text{p}, \hat{y}_\text{cept}^\text{p}\right) + \text{BCE}\left(y_\text{cept}^\text{s}, \hat{y}_\text{cept}^\text{s}\right)
    \vspace{-1.5mm}
\end{equation}
where $\text{BCE}\left(\cdot\right)$ is the binary cross-entropy loss. $y_\text{cept}^\text{p}$ and $y_\text{cept}^\text{s}$ are multi-hot labels indicating semantic concepts contained by the paragraph and sentence queries, respectively. $\hat{y}_\text{cept}^\text{s}$ and $\hat{y}_\text{cept}^\text{p}$ are the concept predictions obtained by dot-product between the conceptual dictionary features and the textual query features $\mathcal{Z}_\text{q}$ with a sigmoid activation.

\noindent\textbf{Query Decoder.} Inspired by the spiritual ideas of dynamic-anchor DETR decoders~\cite{condetr, dab-detr, dn-detr, dino_detr}, we design a novel query decoder that enables dynamic position-aware decoding of language queries. Specifically, each query decoder layer consists of a self-attention module, a cross-attention module, and a feed-forward network, with dynamically adjustable anchor boxes to indicate the query-specific location information. Initially, the input anchor boxes are set to all zeros, and an MLP is used to estimate a set of seed anchor boxes based on the cross-modal interactions between $\mathcal{M}_\text{aug}$ and $\mathcal{Z}_\text{q}$, i.e., the output features of the first decoder layer $\mathcal{Q}_{\text{aug}}^{(1)}\in \mathbb{R}^{(N+1)\times D}$ are mapped into the seed anchor boxes $\mathcal{A}_{\text{aug}}^{(1)}\in \mathbb{R}^{(N+1)\times 2}$ by the MLP. For the $(i+1)$-th decoder layer, we first convert the box coordinates of the input anchors $\mathcal{A}_{\text{aug}}^{(i)}$ into high-dimensional sinusoidal embeddings $\mathcal{F}_{\text{a}}^{(i)}\in \mathbb{R}^{(N+1)\times D}$ by a sinusoidal function~\cite{attention, dab-detr} and further obtain $\mathcal{H}_{\text{a}}^{(i)}$ by projecting $\mathcal{F}_{\text{a}}^{(i)}$ with an MLP. Afterwards, we conduct the self-attention operation and update the query features in the decoder as follows:
\vspace{-1.5mm}
\begin{equation}
    \small
    \mathcal{Q}_{\text{aug}}^{(i+1)}\leftarrow
    \text{Self-Attn}
    \left\{\begin{aligned}
        Q &= \varphi_{q}^{c}\left(\mathcal{Q}_{\text{aug}}^{(i)}\right) + \varphi_{q}^{p}\left(\mathcal{H}_{\text{a}}^{(i)}\right) \\
        K &= \varphi_{k}^{c}\left(\mathcal{Q}_{\text{aug}}^{(i)}\right) + \varphi_{k}^{p}\left(\mathcal{H}_{\text{a}}^{(i)}\right) \\ 
        V &= \varphi_{v}^{c}\left(\mathcal{Q}_{\text{aug}}^{(i)}\right)
    \end{aligned}\right.
    \vspace{-1.5mm}
\end{equation}
where $\mathcal{Q}_{\text{aug}}^{(i+1)}$ is the updated query features after the self-attention operation in the $(i+1)$-th decoder layer. The above series of $\varphi$ functions are used to indicate different linear projection layers for the content part or position part of the queries, keys, or values in the self-attention mechanism. Then, we conduct a cross-attention operation to extract useful cross-modal interactive information as follows:
\vspace{-1mm}
\begin{equation}
    \small
    \mathcal{Q}_{\text{aug}}^{(i+1)}\leftarrow
    \text{Cross-Attn}
    \left\{\begin{aligned}
        Q &= \left[\varphi_{q}^{c}\left(\mathcal{Q}_{\text{aug}}^{(i+1)}\right); \varphi_{q}^{p}\left(\mathcal{F}_{\text{a}}^{(i)}\right)\right] \\
        K &= \left[\varphi_{k}^{c}\left(\mathcal{M}_{\text{aug}}\right); \varphi_{k}^{p}\left(\mathcal{F}_{\text{pe}}^\mathcal{M}\right)\right] \\
        V &= \varphi_{v}^{c}\left(\mathcal{M}_{\text{aug}}\right) + \varphi_{v}^{p}\left(\mathcal{F}_{\text{pe}}^\mathcal{M}\right)
    \end{aligned}\right.
    \vspace{-1mm}
\end{equation}
where $\mathcal{F}_{\text{pe}}^\mathcal{M} = \text{PE}\left(\mathcal{M}_\text{aug}\right)$ and the query features $\mathcal{Q}_{\text{aug}}^{(i+1)}$ is then further updated by a feed-forward network as the feature output of the $(i+1)$-th decoder layer. Then the anchor boxes are dynamically updated as $\mathcal{A}_{\text{aug}}^{(i+1)} \leftarrow \mathcal{A}_{\text{aug}}^{(i)} + \Delta\mathcal{A}_{\text{aug}}^{(i)}$, where we utilize an MLP layer to predict the relative offsets, i.e., $\Delta\mathcal{A}_{\text{aug}}^{(i)}\in \mathbb{R}^{(N+1)\times 2}$, based on the updated query features $\mathcal{Q}_{\text{aug}}^{(i+1)}$. $\mathcal{A}_{\text{aug}}^{(i+1)}$ continues to be forwarded to the next decoder layer for computing $\mathcal{F}_{\text{a}}^{(i+1)}$ and $\mathcal{H}_{\text{a}}^{(i+1)}$.

\noindent\textbf{Boundary Prediction.} Based on the output features and attention weights of the last query decoder layer, we simply use an MLP predictor to predict the paragraph timestamps, i.e., $\widehat{\mathcal{T}}_\text{aug}^\text{p}=(\hat{\tau}_{\text{aug}}^{\text{st}}, \hat{\tau}_{\text{aug}}^{\text{ed}})$. Similarly, the sentence timestamps $\widehat{\mathcal{T}}_\text{inf}^\text{s}=\left\{(\hat{\tau}_{\text{j}}^{\text{st}}, \hat{\tau}_{\text{j}}^{\text{ed}})\right\}_{j=1}^{N}$ can also be obtained by feeding last-layer output from the inference branch to the same MLP.

\noindent\textbf{Self-Consistent Boundary Regression.} We improve the attention-agnostic regression loss $\mathcal{L}_\text{reg}$ to make it aware of the model's self-consistent scores, where the main idea is to selectively optimize the regression loss of self-consistent samples for better weakly-supervised regression learning. Self-consistent samples have high attention weights over the pseudo ground-truth intervals, which are more suitable for learning less noisy supervision for accurate boundary prediction. Specifically, the self-consistent boundary regression loss $\mathcal{L}_\text{screg}$ is defined as:
\vspace{-1mm}
\begin{equation}
    \mathcal{L}_\text{screg} = 
    \begin{cases}
    \mathcal{L}_\text{L1} + \mathcal{L}_\text{GIoU}, \quad \text{if}\quad \space s_\text{att} > \beta \\
    0, \quad \text{otherwise} 
    \end{cases}
    \vspace{-1mm}
\end{equation}
where $\mathcal{L}_{\text{L1}}$ and $\mathcal{L}_{\text{GIoU}}$ respectively represent L1 and Generalized Intersection over Union (GIoU) ~\cite{giou} loss computed between $(\hat{\tau}_{\text{aug}}^{\text{st}}, \hat{\tau}_{\text{aug}}^{\text{ed}})$ and $(\tau_{\text{aug}}^{\text{st}}, \tau_{\text{aug}}^{\text{ed}})$. $\beta$ is set to $0.5$ and $s_\text{att}$ is the attention sum over the pseudo ground-truth interval.

\subsection{Inference Branch}
In our siamese framework, the inference branch shares the same parameter weights and network structure with the augmentation branch, i.e., a video encoder, a query encoder, and a query decoder. The only difference lies in the input streams and objectives, i.e., the inference branch receives a normal video during training to learn in-domain cross-modal correspondence that cannot be acquired through the pseudo video stream, which significantly improves the generalization ability of the model. Specifically, it receives the encoded normal video features $\mathcal{M}_\text{inf}$ and the encoded text query features $\mathcal{Z}_\text{q}$ as the decoder input and generates the hidden query features $\mathcal{Q}_\text{inf}$ for boundary prediction.

\noindent\textbf{Order-guided Attention Loss.} The chronological prior given by the sentence order provides explicit guidance for learning cross-modal alignment between the video and language features during decoding. To learn the order-guided cross-modal correspondence, we constrain cross-modal attention weights in the decoder as follows:
\vspace{-3mm}
\begin{equation}
    \small
    \mathcal{L}_\text{oga} = max\left(0, \Delta m T + \sum_{t=1}^{T}t\alpha_\text{s}^{j}(t) - \sum_{t=1}^{T}t\alpha_\text{s}^{j+1}(t)\right)
    \vspace{-3mm}
    \label{eq: oga}
\end{equation}
where $\alpha_\text{s}^{j}(t)$ and $\alpha_\text{s}^{j+1}(t)$ are attention weights over video features for the $j$-th and $(j+1)$-th sentence from the last decoder layer, respectively. $\Delta m$ is the minimal distance between attention centroids. This loss is mainly contributed by the inference branch with $\Delta m=\frac{1}{2N}$ and partly contributed by the augmentation branch with $\Delta m=\frac{1}{4N}$.

\noindent\textbf{Auxiliary Losses.} To exploit more guidance for weakly-supervised representation learning, we employ three auxiliary losses including a cross-branch loss $\mathcal{L}_\text{cb}$, an anchor ranking loss $\mathcal{L}_\text{ar}$ and a pseudo attention loss $\mathcal{L}_\text{pa}$. Specifically, $\mathcal{L}_\text{cb}$ utilizes the semantic consistency constraint of output features across siamese branches, which is calculated analogous to MoCo~\cite{moco}. $\mathcal{L}_\text{ar}$ is used for inducing the decoder to learn a set of order-preserving anchor boxes, which is computed on the anchor boxes like in the equation~(\ref{eq: oga}). $\mathcal{L}_\text{pa}$ makes use of attention supervision between the pseudo video and language features from the augmentation branch, which follows the calculation proposed in ~\cite{ablr, prvg}.

\subsection{Training and Inference}
\noindent\textbf{Weakly-Supervised Learning.} The weakly-supervised loss of our proposed framework can be formulated as a weighted sum of the two losses from the siamese branches as $\mathcal{L}_\text{WS} = \lambda_\text{screg}\mathcal{L}_\text{screg} + \lambda_\text{oga}\mathcal{L}_\text{oga}$, where $\lambda_\text{screg}$ and $\lambda_\text{oga}$ are scalar weights to balance the contributions of the two different losses. The overall loss for the weakly-supervised model is defined as $\mathcal{L} = \mathcal{L}_\text{WS} + \lambda_\text{csc}\mathcal{L}_\text{csc} + \mathcal{L}_\text{aux}$.

\noindent\textbf{Semi-Supervised Learning.} Although our framework is initially designed for weakly-supervised learning, it can be easily adapted for end-to-end semi-supervised learning. Specifically, in addition to $\mathcal{L}_{\text{WS}}$ which is calculated on all training samples, we only need to employ an extra fully-supervised loss $\mathcal{L}_{\text{FS}}$ on those fully-annotated samples without changing any part of the network structure. The semi-supervised loss $\mathcal{L}_{\text{SS}}$ is defined as $\mathcal{L}_{\text{SS}} = \mathcal{L}_{\text{WS}} + \mathcal{L}_{\text{FS}}$, where $\mathcal{L}_{\text{FS}}$ consists of a regression loss and an attention loss and is calculated on the labeled samples in the inference branch.

\noindent\textbf{Model Inference.} As mentioned, the augmentation branch and conceptual semantic connector are discarded, while other inference branch modules are preserved for testing.
\section{Experiment}
\label{sec:experiment}
\subsection{Datasets and Metrics}
\noindent\textbf{ActivityNet-Captions.} ActivityNet-Captions~\cite{activitynet_captions} dataset is a large-scale dataset with diverse open-domain content sourced from ActivityNet dataset~\cite{activitynet_captions}. There are 14,926 videos and 19,811 localized video-paragraph pairs in total. Each video lasts for 117.60 seconds and each paragraph consists of 3.63 sentences on average. The entire dataset is divided into train/val\_1/val\_2 sets containing 10,009/4,917/4,885 video-paragraph pairs, respectively. We follow prior works~\cite{dense_grounding, svptr} to use val\_2 set for testing.

\noindent\textbf{Charades-CD-OOD.} Charades-STA dataset~\cite{tall} is built from the Charades dataset~\cite{charades_ori} with indoor activities. Following the previous work~\cite{svptr}, we adopt a reorganized version of Charades-STA named Charades-CD-OOD proposed in ~\cite{closerlook}. It is divided into train/val/test\_ood sets consisting of 4,564/333/1,440 video-paragraph pairs, respectively. Specifically, the average video duration is 30.78 seconds and the average paragraph length is 2.41 sentences.

\noindent\textbf{TACoS.} TACoS dataset~\cite{tacos} is constructed from the MPII corpus~\cite{MPII} tailored for cooking activities and kitchen scenarios. There are 127 videos in total with each video paired with multiple paragraphs at different granularities. Concretely, there are 1,107, 418, and 380 video-paragraph pairs for training, validation, and testing, respectively. The average video length and number of sentences in the paragraph are 4.79 minutes and 8.75 in this dataset, respectively.

\noindent\textbf{Evaluation Metrics.} Following previous works~\cite{dense_grounding, svptr}, we adopt mean Intersection over Union (i.e., mIoU) and recall under IoU threshold of $m$ (i.e., $\text{R@m}$) as our evaluation metrics. The metrics are averaged over all sentences and $m$ is set to be $\{0.3, 0.5, 0.7\}$ for ActivityNet-Captions and Charades-CD-OOD, and $\{0.1, 0.3, 0.5\}$ for TACoS.
 
\subsection{Implementation Details}
For fair comparison with existing works~\cite{dense_grounding, prvg}, we adopt the same C3D network~\cite{c3d} and Glove model~\cite{glove} as feature extractors. The number of sampled video clips $T$ is set to be $256$, $128$, and $512$ for ActivityNet-Captions, Charades-CD-OOD, and TACoS datasets, respectively. We train the model using Adam~\cite{adam} optimizer with a fixed learning rate of $0.0001$ and a batch size of $32$, $32$, and $16$ for ActivityNet-Captions, Charades-CD-OOD and TACoS, respectively. We select top-100 high-frequency concepts for each dataset, and the loss weights $\left\{\lambda_\text{screg}, \lambda_\text{oga}, \lambda_\text{csc}\right\}$ are set to $\left\{2, 1, 10\right\}$. The number of encoder and decoder layers is set to be $3$, and the hidden size $D$ is $256$ in all settings.

\subsection{Comparison with State-of-the-arts}
We compare the proposed SiamGTR with existing state-of-the-art methods for VPG to demonstrate the superiority of our framework. Specifically, 3D-TPN~\cite{2dtan, dense_grounding}, DepNet~\cite{dense_grounding}, PRVG~\cite{prvg}, SVPTR~\cite{svptr} and HSCNet~\cite{hscnet} are fully-supervised approaches requiring temporal annotations for the entire dataset. Besides, the semi-supervised setting has been studied in ~\cite{svptr} with several methods developed. For fair comparison with our method, we regard the reconstruction learning method WSSL~\cite{wsvg_recon_1} as one baseline and further develop a more competitive model called Weakly-Supervised Temporal Paragraph Network (WSTPN) by incorporating Beam Search~\cite{beam} into WSTAN~\cite{wstan}. Specifically, WSTPN utilizes a complete paragraph for multiple instance learning and searches the best sequence of timestamps with the highest overall confidence while maintaining a consistent temporal order with the input sentences.

Comprehensive results over three different datasets are shown in Table~\ref{tab: comp_activity}, Table~\ref{tab: comp_charades}, and Table~\ref{tab: comp_tacos}, where FS/SS/WS are used to indicate fully/semi/weakly-supervised settings of video paragraph grounding, respectively. First of all, our SiamGTR remarkably surpasses all the other methods under the same supervision in all metrics over the three datasets. Concretely, our framework outperforms WSTPN by $13.78\%$, $3.72\%$, and $11.16\%$ in mIoU on ActivityNet-Captions, Charades-CD-OOD and TACoS datasets, respectively. Compared to semi-supervised methods using a considerable number of temporal labels, our weakly-supervised method is also able to achieve comparable or even better results, which demonstrates the effectiveness of our framework in efficient weakly-supervised learning. Furthermore, our framework is flexible and can be easily adapted to semi-supervised learning for further gains. As shown, our semi-supervised model outstrips all semi-supervised state-of-the-arts by a large margin, and it performs better or on par with the fully-supervised SVPTR on all three datasets.

\begin{table}[t]
    \centering
    \footnotesize
    \caption{Comparison on ActivityNet-Captions dataset.}
    \vspace{-3mm}
    \begin{tabular}{l|ccccc}
    \toprule[1pt]
    Method & Setting & R@0.3 & R@0.5 & R@0.7 & mIoU \\
    \midrule[1pt]
    3D-TPN~\cite{2dtan} & FS & 67.56 & 51.49 & 30.92 & - \\
    DepNet~\cite{dense_grounding} & FS & 72.81 & 55.91 & 33.46 & - \\
    PRVG~\cite{prvg} & FS & 78.27 & 61.15 & 37.83 & 55.62 \\
    SVPTR~\cite{svptr} & FS & 78.07 & 61.70 & 38.36 & 55.91 \\
    HSCNet~\cite{hscnet} & FS & 81.89 & 66.57 & 44.03 & 59.71 \\
    \midrule[1pt]
    DepNet~\cite{dense_grounding} & SS & 61.46 & 45.14 & 26.78 & 44.11 \\
    VPTR~\cite{svptr} & SS & 72.80 & 53.14 & 29.07 & 50.08 \\
    SVPTR~\cite{svptr} & SS & 73.39 & 56.72 & 32.78 & 51.98 \\
    \rowcolor{gray!38}SiamGTR (Ours) & SS & \color{black}\textbf{78.75} & \color{black}\textbf{59.11} & \color{black}\textbf{34.12} & \color{black}\textbf{54.57} \\
    \midrule[1pt]
    WSSL~\cite{wsvg_recon_1} & WS & 41.98 & 23.34 & - & 28.33 \\
    WSTPN~\cite{wstan} & WS & 57.74 & 33.02 & 13.62 & 38.54 \\
    \rowcolor{gray!38}SiamGTR (Ours) & WS & \color{black}\textbf{75.43} & \color{black}\textbf{57.23} & \color{black}\textbf{30.56} & \color{black}\textbf{52.32} \\
    \midrule[1pt]
    \end{tabular}
    \label{tab: comp_activity}
    \vspace{-5mm}
\end{table}
\begin{table}[t]
    \centering
    \footnotesize
    \caption{Comparison on Charades-CD-OOD dataset.}
    \vspace{-3mm}
    \begin{tabular}{l|ccccc}
    \toprule[1pt]
    Method & Setting & R@0.3 & R@0.5 & R@0.7 & mIoU \\
    \midrule[1pt]
    DepNet~\cite{dense_grounding} & FS & 45.61 & 27.59 & 10.69 & 29.30 \\
    STLG~\cite{stlg} & FS & 48.30 & 30.39 & 9.79 & - \\
    SVPTR~\cite{svptr} & FS & 55.14 & 32.44 & 15.53 & 36.01 \\
    \midrule[1pt]
    DepNet~\cite{dense_grounding} & SS & 43.03 & 25.07 & 10.14 & 28.09 \\
    STLG~\cite{stlg} & SS & 46.15 & 29.43 & 9.38 & - \\
    VPTR~\cite{svptr} & SS & 45.13 & 24.98 & 10.22 & 28.92 \\
    SVPTR~\cite{svptr} & SS & 50.31 & 28.50 & 12.27 & 32.13 \\
    \rowcolor{gray!38}SiamGTR (Ours) & SS & \color{black}\textbf{59.07} & \color{black}\textbf{35.47} & \color{black}\textbf{14.95} & \color{black}\textbf{38.87} \\
    \midrule[1pt]
    WSSL~\cite{wsvg_recon_1} & WS & 35.86 & 23.67 & 8.27 & - \\
    WSTPN~\cite{wstan} & WS & 48.61 & 29.27 & 10.79 & 33.49 \\
    \rowcolor{gray!38}SiamGTR (Ours) & WS & \color{black}\textbf{57.33} & \color{black}\textbf{33.87} & \color{black}\textbf{12.31} & \color{black}\textbf{37.21} \\
    \midrule[1pt]
    \end{tabular}
    \label{tab: comp_charades}
    \vspace{-8mm}
\end{table}
\subsection{Ablation Study}
We conduct ablation studies to investigate the contributions of different components on ActivityNet-Captions dataset.

\noindent\textbf{Effectiveness of data augmentation.} To evaluate the influences of the random boundary shifting and random re-sampling operations for data augmentation, we remove one or both of them from the training pipeline and the results are shown in Table~\ref{tab: ablation1} (a) $\sim$ (d). The model performance clearly degrades after the removal, which demonstrates the necessity of increasing sample diversity and alleviating overfitting for weakly-supervised cross-modal regression learning.
\begin{table}[t]
    \centering
    \footnotesize
    \caption{Comparison on TACoS dataset.}
    \vspace{-3mm}
    \begin{tabular}{l|ccccc}
    \toprule[1pt]
    Method & Setting & R@0.1 & R@0.3 & R@0.5 & mIoU \\
    \midrule[1pt]
    3D-TPN~\cite{2dtan} & FS & 55.05 & 40.31 & 26.54 & - \\
    DepNet~\cite{dense_grounding} & FS & 56.10 & 41.34 & 27.16 & - \\
    PRVG~\cite{prvg} & FS & 61.64 & 45.40 & 26.37 & 29.18 \\
    SVPTR~\cite{svptr} & FS & 67.91 & 47.89 & 28.22 & 31.42 \\
    HSCNet~\cite{hscnet} & FS & 76.28 & 59.74 & 42.00 & 40.61 \\
    \midrule[1pt]
    DepNet~\cite{dense_grounding} & SS & 40.27 & 26.95 & 16.54 & 18.68 \\
    VPTR~\cite{svptr} & SS & 61.31 & 40.59 & 21.39 & 26.59 \\
    SVPTR~\cite{svptr} & SS & 63.06 & 40.19 & 20.05 & 26.10 \\
    \rowcolor{gray!38}SiamGTR (Ours) & SS & \color{black}\textbf{67.30} & \color{black}\textbf{49.35} & \color{black}\textbf{31.69} & \color{black}\textbf{32.81} \\
    \midrule
    WSTPN~\cite{wstan} & WS & 28.59 & 10.04 & 4.76 & 9.32 \\
    \rowcolor{gray!38}{SiamGTR (Ours)} & WS & \color{black}\textbf{61.51} & \color{black}\textbf{26.22} & \color{black}\textbf{10.53} & \color{black}\textbf{20.48} \\
    \midrule[1pt]
    \end{tabular}
    \label{tab: comp_tacos}
    \vspace{-5mm}
\end{table}
\begin{table}[t]
    \centering
    \caption{Ablation studies on component designs of our framework. Experimental results are marked from ID (a) $\sim$ (l). RBS and RRS respectively denote the random boundary shifting and random re-sampling operations for pseudo data generation. MPE, CSC and DAB stand for the modulated positional encodings, conceptual semantic connector, and dynamic anchor boxes, respectively.}
    \vspace{-3mm}
    \resizebox{!}{27.5mm}{
    \begin{tabular}{c|cc|ccc|cc}
    \toprule[1pt]
    ID & RBS & RRS & MPE & CSC & DAB & R@0.5 &mIoU \\
    \toprule[1pt]
    (a) & & & \checkmark & \checkmark & \checkmark & \small{47.10} & \small{46.04} \\
    (b) & \checkmark &  & \checkmark & \checkmark & \checkmark & \small{48.19} & \small{46.96} \\
    (c) & & \checkmark & \checkmark & \checkmark & \checkmark & \small{54.24} & \small{50.94} \\
    (d) & \checkmark & \checkmark & \checkmark & \checkmark & \checkmark & \small{57.23} & \small{52.32} \\
    \hline
    (e) & \checkmark & \checkmark &  &  &  & \small{45.25} & \small{44.46} \\
    (f) & \checkmark & \checkmark & \checkmark &  &  & \small{46.47} & \small{45.47} \\
    (g) & \checkmark & \checkmark &  & \checkmark &  & \small{46.96} & \small{45.23} \\
    (h) & \checkmark & \checkmark &  &  & \checkmark & \small{51.34} & \small{48.52} \\
    (i) & \checkmark & \checkmark & \checkmark & \checkmark &  & \small{52.12} & \small{48.75} \\
    (j) & \checkmark & \checkmark & \checkmark &  & \checkmark & \small{53.57} & \small{50.06} \\
    (k) & \checkmark & \checkmark &  & \checkmark & \checkmark & \small{54.09} & \small{50.84} \\
    (l) & \checkmark & \checkmark & \checkmark & \checkmark & \checkmark & \small{57.23} & \small{52.32} \\
    \toprule[1pt]
    \end{tabular}
    }
    \label{tab: ablation1}
    \vspace{-8mm}
\end{table}

\vspace{-4mm}\noindent\textbf{Ablation on module designs.} As shown in Table~\ref{tab: ablation1} (e) $\sim$ (l), we conduct detailed ablation studies on module designs to validate the rationality of the proposed model. As observed, the designs of modulated positional encodings and dynamic anchor boxes in the encoder and decoder are consistently beneficial to improving the model capacity for better performance. The conceptual semantic connector that bridges two types of queries is also effective, and it further boosts the performance by $2.26\%$ in mIoU (50.06\% vs. 52.32\%) even though the network has been equipped with strong encoders and decoders. We notice the dynamic anchors for decoding are the most crucial component given the sharpest performance drop of $3.57\%$ with its removal, which indicates the importance to explicitly represent the intermediate location information for video grounding.
\begin{table}[t]
    \centering
    \caption{Ablation studies on different weakly-supervised losses.}
    \vspace{-3mm}
    \resizebox{!}{11mm}{
    \begin{tabular}{c|cc|cccc}
    \toprule[1pt]
    ID & $\mathcal{L}_\text{screg}$ & $\mathcal{L}_\text{oga}$ & R@0.3 & R@0.5 & R@0.7 & mIoU \\
    \toprule[1pt]
    (a) & & & \small{45.85} & \small{29.82} & \small{10.93} & \small{30.97} \\
    (b) & \checkmark & & \small{46.90} & \small{29.38} & \small{12.06} & \small{31.82} \\
    (c) & & \checkmark & \small{73.58} & \small{54.58} & \small{28.82} & \small{50.62} \\
    (d) & \checkmark & \checkmark & \small{75.43} & \small{57.23} & \small{30.56} & \small{52.32} \\
    \toprule[1pt]
    \end{tabular}
    }
    \label{tab: ablation2}
    \vspace{-4mm}
\end{table}
\begin{table}[t]
    \centering
    \caption{Impact of different auxiliary losses.}
    \vspace{-3mm}
    \resizebox{!}{11.5mm}{
    \begin{tabular}{l|cccc}
    \toprule[1pt]
    Method & R@0.3 & R@0.5 & R@0.7 & mIoU \\
    \toprule[1pt]
    \small{w/o $\mathcal{L}_\text{cb}$} & \small{71.49} & \small{49.78} & \small{24.64} & \small{48.07} \\
    \small{w/o $\mathcal{L}_\text{ar}$} & \small{72.96} & \small{50.72} & \small{26.42} & \small{49.14} \\
    \small{w/o $\mathcal{L}_\text{pa}$} & \small{74.34} & \small{56.01} & \small{30.29} & \small{51.76} \\
    \small{Full Model} & \small{75.43} & \small{57.23} & \small{30.56} & \small{52.32} \\
    \toprule[1pt]
    \end{tabular}
    }
    \label{tab: ablation3}
    \vspace{-4mm}
\end{table}
\begin{table}[t]
    \centering
    \caption{Evaluation on different types of paragraph representation.}
    \vspace{-3mm}
    \resizebox{!}{11.5mm}{
    \begin{tabular}{l|cccc}
    \toprule[1pt]
    Method & R@0.3 & R@0.5 & R@0.7 & mIoU \\
    \toprule[1pt]
    \small{Max-Pooling} & \small{72.61} & \small{52.61} & \small{27.03} & \small{49.55} \\
    \small{Mean-Pooling} & \small{74.31} & \small{53.13} & \small{28.64} & \small{50.57} \\
    \small{Word Concat} & \small{73.79} & \small{52.27} & \small{27.30} & \small{49.96} \\
    \small{Learnable} & \small{75.43} & \small{57.23} & \small{30.56} & \small{52.32} \\
    \toprule[1pt]
    \end{tabular}
    }
    \label{tab: ablation4}
    \vspace{-7.5mm}
\end{table}

\noindent\textbf{Analysis on weakly-supervised losses.} Ablation studies on contributions of two weakly-supervised losses $\mathcal{L}_\text{screg}$ and $\mathcal{L}_\text{oga}$ are shown in Table~\ref{tab: ablation2}. In experiment (a) and (c), we remove $\mathcal{L}_\text{screg}$ but use a plain regression loss $\mathcal{L}_\text{reg}$ for ablation analysis. Firstly, simply employing regressive supervision attains inferior performance because the feature-level cross-modal correspondence can hardly be learned with coordinate-level supervision. Furthermore, we find $\mathcal{L}_\text{oga}$ is critical for learning precise temporal localization since it explicitly guides the model to align video features and language features that are highly likely to be correlated. Besides, we observe that using $\mathcal{L}_\text{screg}$ for selecting high-quality regression samples always brings gains to the performance.

\noindent\textbf{Impact of auxiliary losses.} We investigate the influences of auxiliary losses on the model performance, which involve $\mathcal{L}_\text{cb}$ for exploiting the cross-branch knowledge, $\mathcal{L}_\text{ar}$ as guidance to learn a  set of order-preserving anchor boxes and $\mathcal{L}_\text{pa}$ to make use of the feature alignment supervision from the augmentation branch. As presented in Table~\ref{tab: ablation3}, all three auxiliary losses bring positive impacts, with $\mathcal{L}_\text{cb}$ being the most effective to improve the mIoU metric by $4.25\%$. The reason might lie in the rich complementary knowledge and consistency supervision across the siamese branches.

\noindent\textbf{Impact of the paragraph representation.} The comparison of different types of paragraph representation is shown in Table~\ref{tab: ablation4}. Four different schemes are included, i.e., mean-pooling or max-pooling the sentence features, embedding the sequence of all word tokens in the paragraph, and using a learnable query token to extract global information. It is clear that adaptively learning a paragraph representation with our method achieves the best performance and the max-pooling scheme performs the worst because it loses too much detailed information. The mean-pooling scheme performs slightly better than the word-concat scheme, which may attribute to the advantage of late fusion at feature level.

\subsection{Visualization}
\begin{figure}[t]
    \centering
    \includegraphics[width=1\linewidth]{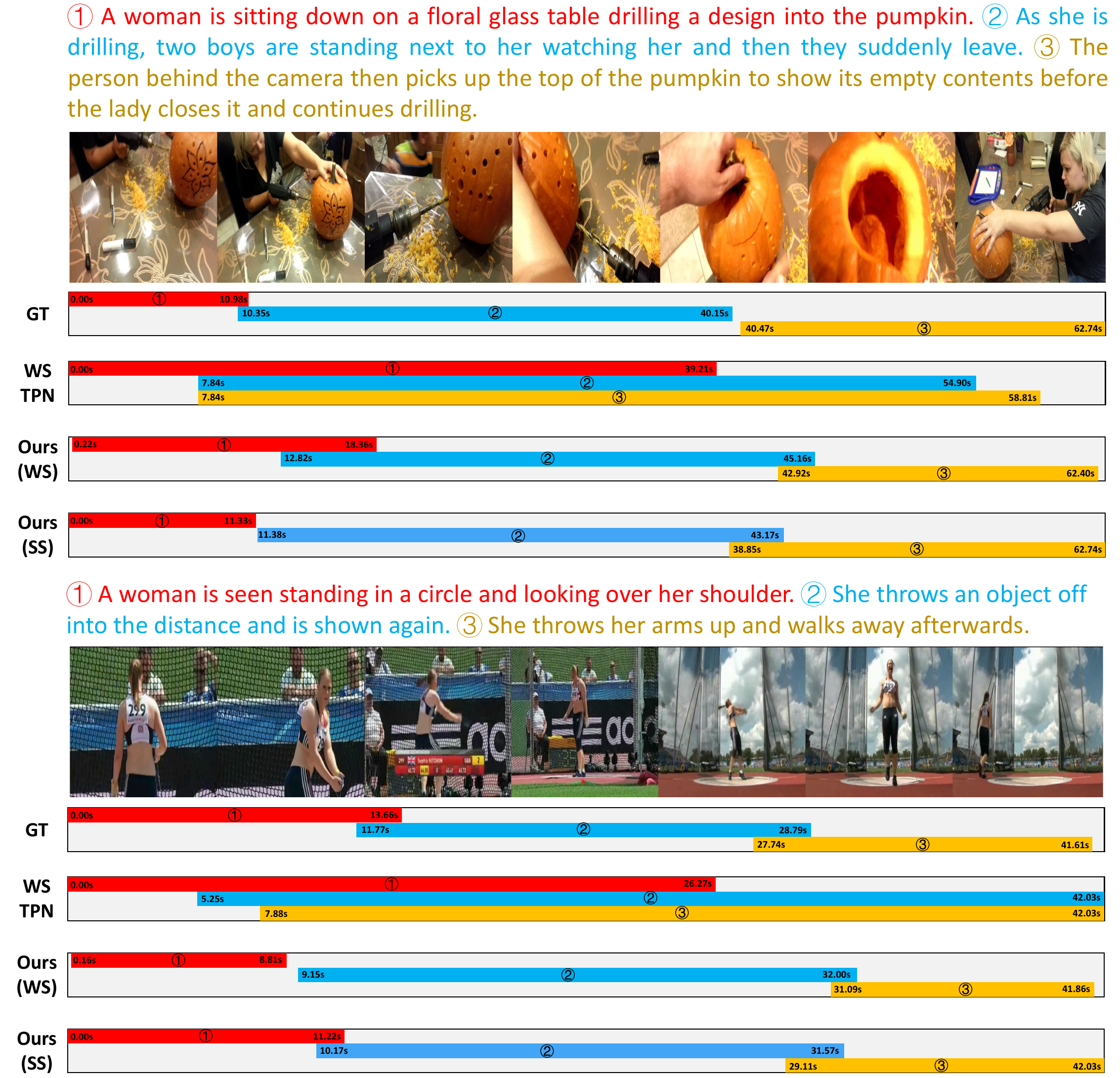}
    \vspace{-6mm}
    \caption{Visualization of prediction results from different models.}
    \label{fig: visualize}
    \vspace{-5mm}
\end{figure}
In Figure~\ref{fig: visualize}, we intuitively visualize the predicted timestamps from WSTPN and our proposed model trained under the weakly-supervised or semi-supervised setting. Overall, the predicted timestamps given by WSTPN coarsely capture the sentence order relations but with inaccurate boundaries. In contrast, our weakly-supervised model achieves much better results. It is notable that our semi-supervised model generates more fine-grained boundaries, which shows the advantage of our siamese framework in jointly leveraging fewer and weaker labels for efficient learning.

\vspace{-1mm}
\section{Conclusion}
\label{sec:conclusion}
In this work, we explore the weakly-supervised setting in video paragraph grounding (i.e., WSVPG) to eliminate the dependence of the temporal annotations. To achieve this goal, we propose a novel siamese learning framework to jointly learn the cross-modal feature alignment and temporal coordinate regression without ground-truth supervision. Specifically, we design a novel Siamese Grounding TRansformer (SiamGTR) consisting of an augmentation branch and an inference branch. The augmentation branch utilizes the boundary supervision provided by temporally regressing a complete paragraph in a pseudo video, and the inference branch learns the order-guided cross-modal correspondence of multiple sentences in a normal video. Extensive experiments verify the effectiveness of our framework.

{\footnotesize{\noindent\textbf{Acknowledgements.}
This work was supported partially by the NSFC (U21A20471, U22A2095, 62076260, 61772570), Guangdong Natural Science Funds Project (2020B1515120085, 2023B1515040025), Guangdong NSF for Distinguished Young Scholar (2022B1515020009), and Guangzhou Science and Technology Plan Project (202201011134).}}

\clearpage
{
    \small
    \bibliographystyle{ieeenat_fullname}
    \bibliography{main}
}

\end{document}